\documentclass[letterpaper, 10 pt, conference]{ieeeconf}  % Comment this line out if you need a4paper

\IEEEoverridecommandlockouts                              % This command is only needed if 
\usepackage{multicol}
\usepackage{caption}
\usepackage[bookmarks=true]{hyperref}
\usepackage{tabularx}
\usepackage[latin1]{inputenc} 
\usepackage[cmex10]{amsmath}
\usepackage{amsfonts} 
\usepackage{amssymb} 
\usepackage{amsbsy}
\usepackage{float}
\usepackage{hyperref} 
\usepackage[scriptsize,tight]{subfigure}
\usepackage{algpseudocode}
\usepackage{comment}
\usepackage{bigstrut}
\usepackage[]{algorithm}
\usepackage{times} % assumes new font selection scheme installed
\usepackage[pdftex]{graphicx}
\graphicspath{{./figures/}}
\usepackage[disable]{todonotes}

\newcommand{\VEC}[1]{\mathbf{#1}}
\newcommand{\symVEC}[1]{\boldsymbol{#1}}
\newcommand{\dVEC}[1]{\dot{\mathbf{#1}}}
\newcommand{\ddVEC}[1]{\ddot{\mathbf{#1}}}
\newcommand{\dsymVEC}[1]{\dot{\boldsymbol{#1}}}

\newcommand*\Bell{\ensuremath{\boldsymbol\ell}}
\newcommand*\dBell{\ensuremath{\boldsymbol{\dot{\ell}}}}
\newcommand{\eqsref}[2]{(\ref{#1}-\ref{#2})}

\renewcommand{\baselinestretch}{0.99}

\title{\LARGE \bf
Structured contact force optimization for kino-dynamic motion generation
}

\author{Alexander Herzog$^{1}$, Stefan Schaal$^{1,2}$, Ludovic Righetti$^{1}$% <-this % stops a space
\thanks{\scriptsize This research was mainly supported by the
    Max-Planck-Society, the Max Planck ETH Center for Learning Systems and the European Research Council (ERC) under
    the European Union's Horizon 2020 research and innovation
    programme (grant agreement No 637935). It was also supported by
    by National Science Foundation grants IIS-1205249, IIS-1017134, EECS-0926052, the Office of Naval Research and the Okawa Foundation.}
    \thanks{\scriptsize$^{1}$Autonomous Motion Department, Max Planck
    Institute for Intelligent Systems, %
    T\"ubingen, Germany. {\tt\small first.lastname@tuebingen.mpg.de}}%
\thanks{\scriptsize$^{2}$CLMC Lab, University of Southern
    California, Los Angeles, USA.}%
}

\begin{document}

\maketitle
\thispagestyle{empty}
\pagestyle{empty}

%%%%%%%%%%%%%%%%%%%%%%%%%%%%%%%%%%%%%%%%%%%%%%%%%%%%%%%%%%%%%%%%%%%%%%%%%%%%%%%%

\begin{abstract}
Optimal control approaches in combination with trajectory optimization have recently proven to be a promising control strategy for legged robots. Computationally efficient and robust algorithms were derived using simplified models of the contact interaction between robot and environment such as the linear inverted pendulum model (LIPM). However, as humanoid robots enter more complex environments, less restrictive models become increasingly important. As we leave the regime of linear models, we need to build dedicated solvers that can compute interaction forces together with consistent kinematic plans for the whole-body. In this paper, we address the problem of planning robot motion and interaction forces for legged robots given predefined contact surfaces. The motion generation process is decomposed into two alternating parts computing force and motion plans in coherence. We focus on the properties of the momentum computation leading to sparse optimal control formulations to be exploited by a dedicated solver. In our experiments, we demonstrate that our motion generation algorithm computes consistent contact forces and joint trajectories for our humanoid robot. We also demonstrate the favorable time complexity due to our formulation and composition of the momentum equations.
\end{abstract}

%%%%%%%%%%%%%%%%%%%%%%%%%%%%%%%%%%%%%%%%%%%%%%%%%%%%%%%%%%%%%%%%%%%%%%%%%%%%%%%%
\section{Introduction}
Trajectory optimization and optimal control approaches have recently been very successful
to control the locomotion of legged robots. Simplified linear models of the dynamics,
usually variations of the linear inverted pendulum model \cite{Kajita:2003uh},\cite{Audren:2014gl},\cite{Englsberger:2015jp},\cite{Herdt:2010bh} have been widely used
to compute the center of mass motion of legged robots, in particular biped robots.
The linearity can be exploited to formulate optimal control problems that can be solved
with quadratic programs, therefore allowing for fast computations of solutions in
a model predictive control manner.

While these simplified models are useful, generalization to more complex situations where multiple non
co-planar contact points might be desirable is limited. Moreover, they do not take into account
how the generation of angular momentum affects the behavior of the robot, nor how individual
contact points should be controlled to create a desired motion.

On the other hand, several contributions have shown that using the full
dynamics model of the robot could be beneficial to generate even more complex behaviors \cite{Dai:2014tp},\cite{Lengagne:2013fq}.
They have the advantage of explicitly taking into account the contact interactions with the
environment and the dynamics of the robot.
The problem with such approaches is that they usually require to solve non convex optimization
problems which are computationally demanding. Several approaches have addressed the problem
using variations of differential dynamic programming to compute optimal trajectories. The drawback
of such algorithms is that they do not allow to easily add constraints on the state and controls.
Other approaches formulate the problem as non-convex optimization problems, however in these cases
the solvers used are generally off the shelf nonlinear solvers that do not exploit the structure of 
problem.

This paper addresses the problem of planning robot motion and interaction forces of a 
legged robot given a set of predefined contact points. In \cite{Herzog:2015}, the authors showed that such
an approach could be taken to compute with a mild computational complexity trajectories
that could be then directly included in whole-body control approaches and executed
on a simulated humanoid.
With a similar approach \cite{Carpentier:2016}, interesting behaviors were demonstrated on a robot, confirming the interest in resolving such mathematical problems more efficiently. 

In this paper, we analyze the structure of the problem induced by the physical model
inherent of legged robots modeled through rigid body dynamics. In particular, we decompose
the planning problem into two alternating optimization phases: first the contact forces
and overall robot momentum are computed to satisfy dynamic constraints and in a second
step the kinematic plan is resolved to satisfy the dynamic plan. Alternating these two phases
lead to a locally optimal solution for the full kinodynamic plan.

We concentrate our efforts on the momentum optimal control problem. First, we show
that by choosing the right representation of contact forces and momentum we can rewrite
the problem as a Quadratically Constrained Quadratic Program for which a convex approximation
of the constraints can be explicitly (and trivially) constructed. We then present two
formulations, using either a simultaneous or sequential optimal control formulation.
We show how the sequential formulation of the problem can 
reduce the size of the problem while preserving its sparse structure for more efficient computation.
Finally, preliminary numerical experiments demonstrate that exploiting this structure in a dedicated solver
can significantly improve computational efficiency and quality of the solutions when compared
to the same problem solved with a state of the art general purpose nonlinear solver.

The remainder of this paper is structured as follows. In Sec. \ref{sec:motion_generation} we propose a motion generation algorithm that is based on optimization of the momentum equations. We write out different representations of contact forces in Sec. \ref{sec:force_rep} and discuss their algebraic properties. These properties are then exploited in Sec. \ref{sec:opt_ctrl_formulations} to construct different variations of optimal control problems on the momentum equations of a floating-base robot. Then, in Sec. \ref{sec:experiments} we show experiments with the proposed motion generation algorithm and conclude the paper.

%
% Problem Formulation
%
\section{Kinodynamic Motion Generation}\label{sec:motion_generation}

We are interested in the problem of motion generation for a
robot in the presence of contact forces. In this section, we are going to write out an algorithm based on an optimal control formulation that is suited for motion generation in floating-base robots. Our underlying dynamics model
is

\begin{align}\label{eq:eom}
  \VEC{M}(\VEC{q})\ddVEC{q} + \VEC{N}(\VEC{q}, \dVEC{q}) =
  \VEC{S}^T \symVEC{\tau}_q + \VEC{J}_{e}^T\symVEC{\lambda},
\end{align}

where $\VEC{q}$ is the joint configuration, $\VEC{M}, \VEC{N}$ the inertia matrix and nonlinear terms, $\symVEC{\tau}_q$ are the actuation torques, $\VEC{S}^T$ the selection matrix, $\VEC{J}_e$ the end effector Jacobian and $\symVEC{\lambda} = [\dots \VEC{f}_e^T~ \symVEC{\tau}_e^T \dots]^T$
is the vector of forces $\VEC{f}_e$ and torques $\symVEC{\tau}_e$
acting at end effector $e$. The equations of motion of a
floating-base robot consist of two parts, the manipulator dynamics
describing the torque at each joint, and the 6 rows of Newton-Euler
equations that describe the change of the overall momentum in the
system. The dynamics can be separated into the 6 rows of Newton-Euler
equations and the actuated part as follows

  \begin{align} \label{eq:eom_decomposed_1} 
    &\VEC{H}(\VEC{q})\ddVEC{q}
    + \dVEC{H}(\VEC{q})\dVEC{q} = \\ \nonumber
    &\begin{bmatrix}
      M\VEC{g} + \sum \VEC{f}_e \\
      \sum \symVEC{\tau}_e + \sum (\VEC{x}_{e}(\VEC{q}) -
      \VEC{x}_{CoM}(\VEC{q})) \times \VEC{f}_e
    \end{bmatrix} \\
   \label{eq:eom_decomposed_2}
    &\bar{\VEC{M}}(\VEC{q})\ddVEC{q} + \bar{\VEC{N}}(\VEC{q}, \dVEC{q})
    = \symVEC{\tau}_q + \bar{\VEC{J}}_{e}^T\symVEC{\lambda},
  \end{align}
  
with the contact location $\VEC{x}_{e}(\VEC{q})$ and center of
mass (CoM) $\VEC{x}_{CoM}(\VEC{q})$ computed through forward
kinematics. $\VEC{H}(\VEC{q})$ is the centroidal momentum matrix that
maps
$\dVEC{q}$ onto the overall linear and angular momentum of the system \cite{Orin:2008ge}. $\bar{\VEC{M}}, \bar{\VEC{N}}, \bar{\VEC{J}}_{e}^T$ represent quantities from Eq. \eqref{eq:eom} corresponding to the actuated joints.\\
From Eqs.~\eqsref{eq:eom_decomposed_1}{eq:eom_decomposed_2}, we observe that the planning
process can be decomposed in two steps. First, a vector of joint
trajectories $\VEC{q}(t)$ is found together with contact force
profiles $\symVEC{\lambda}(t)$ that satisfy
Equation~\eqref{eq:eom_decomposed_1}. Then, suitable torques can be
computed readily from Equation~\eqref{eq:eom_decomposed_2} assuming
that sufficient joint torques can be provided by the actuators. In the
following, we will discuss an optimal control approach that finds
joint trajectories $\VEC{q}(t)$ together with force profiles
$\symVEC{\lambda}(t)$ that satisfy Eqs.~\eqsref{eq:eom_decomposed_1}{eq:eom_decomposed_2}.
Our motion generation algorithm will be phrased as an optimal
control problem of the form

\newcommand*\hofq{\begin{bmatrix}
    \VEC{x}_{CoM}(\VEC{q}_t) \\
    \VEC{H}(\VEC{q}_t)\dVEC{q}_t% + \dVEC{H}(\VEC{q}_t)\dVEC{q}_t
  \end{bmatrix}}
\begin{align}
  \label{eq:kino_dyn_opt}
  \underset{\VEC{q}, \VEC{h}, \symVEC{\lambda}, \VEC{c}}{\text{min.}}
  &&J(\VEC{q}, \VEC{h}, \symVEC{\lambda}, \VEC{c})
  & = \sum_t J_t(\VEC{q}_t) + J'_t(\VEC{h}_t, \symVEC{\lambda}_t, \VEC{c})\\ %|| \VEC{h}_t - \bar{\VEC{h}}_t ||^2 +
%    || \symVEC{\lambda}_t - \bar{\symVEC{\lambda}}_t||^2 \\
  %
  \label{eq:kino_dyn_opt_2}
	\text{s.t.}&& \VEC{c}_{e,t}& = \VEC{x}_{e}(\VEC{q}_t) \\
  \label{eq:kino_dyn_opt_3}
  && \VEC{h}_t
  &=\hofq \\
  \label{eq:kino_dyn_opt_4}
  && \VEC{h}_{t+1}
  &= \VEC{h}_t + \Delta \symVEC{f}_t(\VEC{h}_t, \symVEC{\lambda}_t) \\
  &&\symVEC{\lambda}_t, \VEC{c}_t
  & \in \mathcal{S}_t
\end{align}

where we express objective functions on kinematic quantities $J_t(\VEC{q})$  as well as on momentum and contact locations $J'_t(\VEC{h}_t, \symVEC{\lambda}_t, \VEC{c})$. For instance could $J_t(\VEC{q})$ enforce an end effector motion from one contact location to another. Although contact locations $\VEC{c}_{e,t}$ and momentum $\VEC{h}_t$ can be written as functions of $\VEC{q}$ explicitly, we introduce redundant variables in order to simplify the optimization process as will become clear shortly. Eqs.~\eqsref{eq:kino_dyn_opt_2}{eq:kino_dyn_opt_3} state explicitly how contact locations and momentum relate through forward kinematics. The momentum equations of our robot model \eqref{eq:eom_decomposed_1} are then expressed as constraint in Eq. \eqref{eq:kino_dyn_opt_4} with discretization parameter $\Delta$. $\mathcal{S}$ is
the set of feasible contact forces as will be discussed in more details later.
The use of redundant variables allows us to
decompose the overall problem into two better structured mathematical
programs that allow the application of better informed solvers. The
two sub-problems will be solved iteratively until convergence
resulting in a solution for the original optimization problem. The
first sub-problem is defined as
\begin{align}\label{eq:subprob_momentum}
	\underset{\VEC{h}, \symVEC{\lambda}, \VEC{c}}{\text{min.}} 
	&&&\hspace{-.9cm}\sum_{t,e}
      || \VEC{h}_t - \bar{\VEC{h}}_t ||^2 + 
      || \symVEC{\lambda}_t - \bar{\symVEC{\lambda}}_t||^2 +
      || \VEC{c}_t - \bar{\VEC{c}}_t||^2   \\
	\text{s.t.}
	\label{eq:subproblem_momentum_2}
	&& \VEC{h}_{t+1} &= \VEC{h}_t + \Delta \symVEC{f}_t(\VEC{h}_t, \symVEC{\lambda}_t)  \\
	&&\symVEC{\lambda}_t, \VEC{c}_t & \in \mathcal{S}_t \nonumber
\end{align}

This problem is optimized over the momentum and contact forces only. The resulting
force and momentum profiles are dynamically consistent (i.e. they satisfy dynamics constraints ~\eqref{eq:subproblem_momentum_2}) and are as close as possible to reference profiles $\bar{\VEC{h}} , \bar{\symVEC{\lambda}}, \bar{\VEC{c}}$. However,
the joint state $\VEC{q}$ is ignored in
Problem~\eqref{eq:subprob_momentum}. Objectives on kinematic quantities are considered in the kinematic sub-problem
\begin{align} \label{eq:subproblem_motion}
  \underset{\VEC{q}}{\text{min.}} \sum_t & J_t(\VEC{q}_t) + || \hofq - \bar{\VEC{h}}_t
  ||^2 \\ \nonumber
  &+ || \VEC{x}_e(\VEC{q}_t) - \bar{\VEC{c}}_{e,t} ||^2
\end{align}

where additionally to optimizing $J_t(\VEC{q}_t)$, we put constraints on momentum and contact location (cf. Eqs. \eqsref{eq:kino_dyn_opt_2}{eq:kino_dyn_opt_3}) from the original problem into the cost of our kinematic sub-problem in form of soft constraints. 
We will solve the original Problem~\eqref{eq:kino_dyn_opt} by
iteratively solving
problems~\eqref{eq:subprob_momentum},~\eqref{eq:subproblem_motion} and use the reference trajectories to enforce consistency between the two independent sub-problems.\\

\begin{algorithm}
\caption{Momentum-centric motion generation}\label{algo:motion_generation}
\begin{algorithmic}
\State{initialize $\bar{\VEC{h}}, \bar{\symVEC{\lambda}}, \bar{\VEC{c}}$}
\Repeat
\State{solve Problem \eqref{eq:subproblem_motion}}
\vspace{.1cm}
\State{$\bar{\VEC{h}}, \bar{\VEC{c}_e} := \hofq, \VEC{x}_e(\VEC{q}_t)$}
\vspace{.1cm}
\State{solve Problem \eqref{eq:subprob_momentum}}
\vspace{.1cm}
\State{$\bar{\VEC{h}}, \bar{\VEC{c}_e} := \VEC{h}, \VEC{c}_e$}
\Until{Solution of sub problems \eqref{eq:subproblem_motion},\eqref{eq:subprob_momentum} do not change}
\end{algorithmic}
\end{algorithm}
 
%\begin{itemize}
%\item initialize
%  $\bar{\VEC{h}}, \bar{\symVEC{\lambda}}, \bar{\VEC{c}}$
%\item do
%  \begin{itemize}
%  \item solve Problem~\eqref{eq:subprob_momentum}
%  \item $\bar{\VEC{h}}, \bar{\VEC{c}} := \VEC{h}, \VEC{c}$
%  \item solve Problem~\eqref{eq:subproblem_motion}
%  \item
%    $\bar{\VEC{h}}, \bar{\VEC{c}} := \hofq, \VEC{x}_{e,
%      contact}(\VEC{q}_t)$
%  \end{itemize}
%\item until solutions of subproblems do not change anymore
%\end{itemize}

Each of the sub-problems minimizes parts of the original objective function and satisfies a subset of the original constraints. Additionally, the resulting momentum profiles from one sub-problem are optimized for consistency inside of the other up until convergence.
With the proposed decomposition we have to solve two easier problems,
where Problem~\eqref{eq:subproblem_motion} is an unconstrained
optimization that has no notion of contact forces. Similar formulations, usually without costs on momentum, are often addressed in trajectory optimization for manipulators \cite{Kalakrishnan2011}. Problem~\eqref{eq:subprob_momentum} is a non-convex, but a well structured optimal control
problem as we will discuss in the remainder of this paper.

% 
% Momentum Dynamics Formulations
%
\section{Reaction Force Representations}\label{sec:force_rep}

It is well known that the (linear and angular) momentum of a system can only be
changed by external forces. In this section, we will discuss forces acting at
contact points on a robot structure and we will show how the choice of
representation of contact forces results in different mathematical programs with
beneficial properties to be exploited by optimization algorithms. \\

\subsection{Contact Forces}
A force $\VEC{f}_{e}$ and torque $\symVEC{\tau}_e$ acting at contact
point $\VEC{p}_e$ can be represented equivalently as a force and
torque acting at the CoM $\VEC{r}$ where the torque acting at
$\VEC{r}$ transforms to
\begin{align}
  \label{eq:com_torque_transformation}
	\symVEC{\kappa}_e = \symVEC{\tau}_e + (\VEC{p}_e - \VEC{r}) \times \VEC{f}_e
\end{align}
and $\VEC{f}_e$ remains the same. The change of momentum can then be
expressed equivalently using either of the two representations as
\begin{align}
  \label{eq:momentum_dynamics}
	\dVEC{h}
  &= \begin{bmatrix}
    \dVEC{r} \\ \dBell \\ \dVEC{k}
  \end{bmatrix}
  = \symVEC{f}(\VEC{h}, \VEC{f}, \symVEC{\kappa}) 
  = \begin{bmatrix}
    &\frac{1}{M}\Bell \\
    &M\VEC{g} + \sum_e \alpha_e\VEC{f}_e \nonumber \\
    &\sum_e \alpha_e\symVEC{\kappa}_e
  \end{bmatrix}\\  
  &= \begin{bmatrix}
    &\frac{1}{M}\Bell \\
    &M\VEC{g} + \sum_e \alpha_e\VEC{f}_e \\
    &\sum_e \alpha_e\symVEC{\tau}_e + \sum_e
    \alpha_e(\VEC{p}_e - \VEC{r}) \times \VEC{f}_e
  \end{bmatrix} \\
	\alpha_e &= \begin{cases}
		1,&\text{if $e$ is in contact} \\
		0,& \text{else}
	\end{cases}
\end{align}
where $\Bell, \VEC{k}$ are the linear and angular momentum. From Eq.~\eqref{eq:momentum_dynamics} we can see that depending on how
we chose the point of action the dynamics are either linear or not. One might be tempted to say that the linear dynamics lead to easier mathematical programs, which however is not the case when contact constraints have to be considered.
\begin{figure}
  \centering
  \includegraphics[width=.8\linewidth]{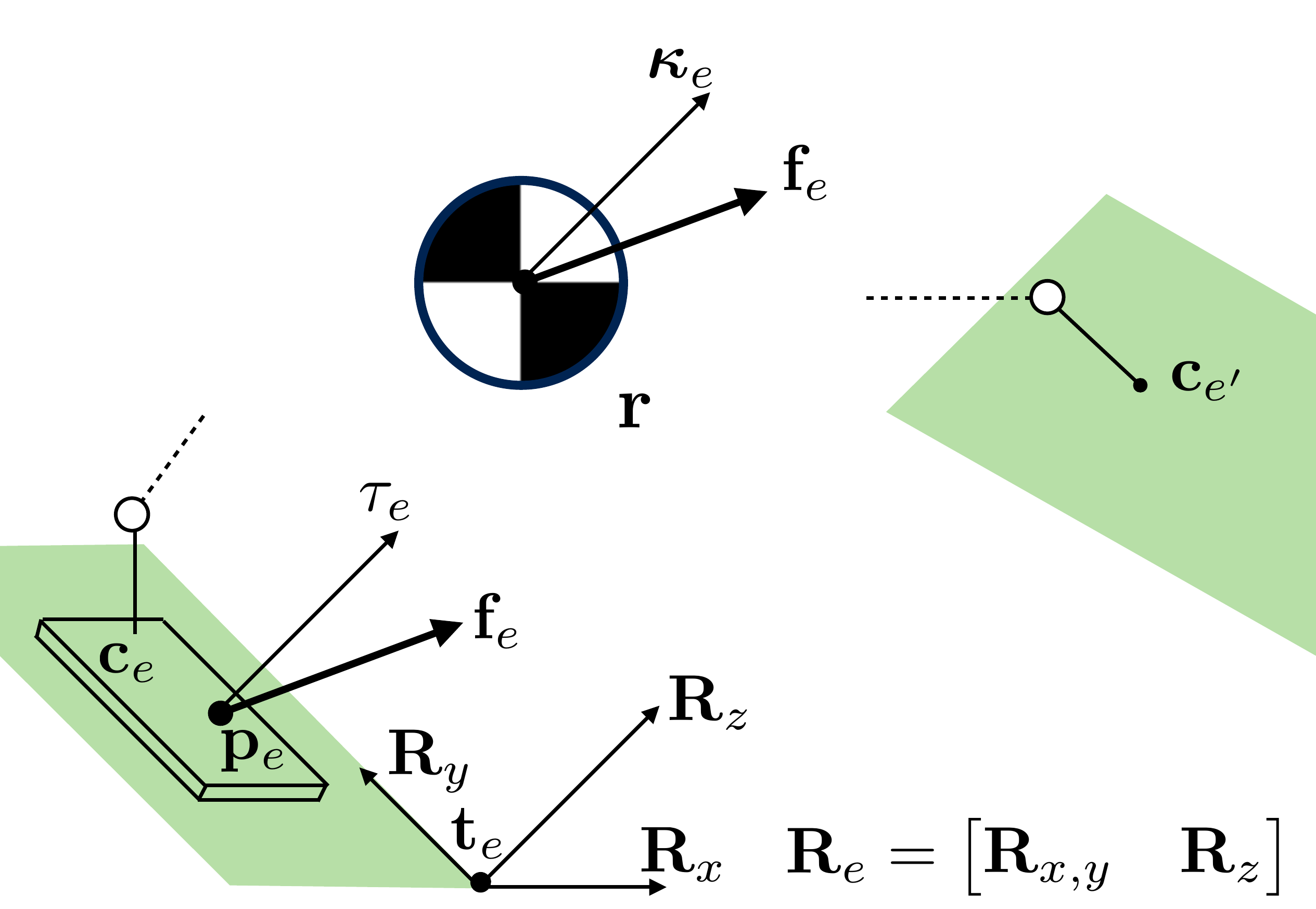}
  \caption{\scriptsize A summary of notation used throughout the discussion of the momentum equations.}\label{fig:contact_forces}
  \vspace{-0.6cm}
\end{figure}
Unilateral contacts, e.g. a hand pushing against a wall
require that $\VEC{f}_e$ remains in a friction cone. On the other
hand, point contacts cannot generate torques, i.e. $\symVEC{\tau}_e = \VEC{0}$. In
both cases, the center of pressure $\VEC{p}_e$ at which $\VEC{f}_e$ is acting
(effectively) needs to reside inside of a support polygon or at the
point of contact $\VEC{c}$. We will now derive constraints on contact forces
that generalize across various types of contacts, such as contact
between surfaces (for instance a flat foot on the ground) or point contacts (for instance an elbow pushing on the table). For the following derivation, we
define a contact surface for each contact $e$ as shown in
Fig.~\ref{fig:contact_forces}. Each contact is located on a surface
described by a rotation matrix
$\VEC{R}_e = \begin{bmatrix} \VEC{R}_{e,x,y}& \VEC{R}_{e,z}\end{bmatrix}$
and translation $\VEC{t}_e$. We can then represent forces, torques and
locations in local coordinates as follows where we drop the index $e$
for better readability
\begin{align} \label{eq:force_rep_transform_1}
	\VEC{f} = \VEC{R} \hat{\VEC{f}},~ 
	\symVEC{\kappa} = \VEC{R} \hat{\symVEC{\kappa}},~ 
	\symVEC{\tau} = \VEC{R}_z \hat{\tau}, \\
	\VEC{p} = \VEC{R}_{x,y} \hat{\VEC{p}} + \VEC{t},~ 
	\VEC{c} = \VEC{R}_{x,y} \hat{\VEC{c}} + \VEC{t}, \\
	 \label{eq:force_rep_transform_3}
	\hat{\tau} \in \mathbb{R},~ \hat{\VEC{p}}, \hat{\VEC{c}} \in \mathbb{R}^2,~ \hat{\VEC{f}}, \hat{\symVEC{\kappa}} \in \mathbb{R}^3
\end{align}
where the hat identifies local representations. We can now write
contact constraints as
\begin{align}
	-\tau_{max} &\leq \hat{\tau} \leq \tau_{max} \label{eq:contact_constraints_1}\\
	-\VEC{p}_{max} &\leq \hat{\VEC{p}} - \hat{\VEC{c}} \leq \VEC{p}_{max} \label{eq:contact_constraints_2}\\
	-\mu \hat{f}_z &\leq \hat{f}_x,\hat{f}_y \leq \mu \hat{f}_z, \label{eq:contact_constraints_3}
\end{align}
where $\mu$ is a friction parameter of the surface and $\VEC{p}_{max}$ is a
vector of support limits and $\tau_{max}$ is a bound on torques that are normal
to the contact surface. If we want to describe a point foot, we set
$\tau_{max} = 0,~ \VEC{p}_{max}=\VEC{0}$. A contact between surfaces, e.g.
floor and foot sole, allow for a wider range of $\hat{\tau}, \hat{\VEC{p}}$.
In Eq.~\eqref{eq:contact_constraints_3} we restrict the force to remain in a
pyramid which is used as an approximation for a friction cone. Note that in
Eq.~\eqref{eq:contact_constraints_2} we consider the support polygon to be
rectangular for a clear notation, however, our derivation holds for more general polyhedra as well. Our contact model can be refined further \cite{Caron:2015} without loss of the structure as derived in the remainder of the paper.
As expressed in Eqs.~\eqsref{eq:contact_constraints_1}
{eq:contact_constraints_3} contact constraints are linear if we express them as functions of
$\VEC{f}_e, \symVEC{\tau}_e$ acting at $\VEC{p}_e$. On the other hand, if we chose to
represent forces and torques as $\VEC{f}_e, \symVEC{\kappa}_e$ acting at
the CoM $\VEC{r}$, we will end up with non-linear contact constraints. In both cases, either the constraints \eqref{eq:contact_constraints_2} become nonlinear or the momentum equations \eqref{eq:momentum_dynamics}  do. In the following,
we will investigate the type of non-linearity that we introduce in
Eqs.~\eqref{eq:momentum_dynamics}, \eqref{eq:contact_constraints_2} and how we can exploit it for development of optimization algorithms.\\

%
% The cross product is a difference of convex quadratic functions
%
\subsection{Algebraic
  categorization} \label{sec:algebraic_categorization}

Depending on the choice of representation of the momentum dynamics
(force acting at $\hat{\VEC{p}}$ or at $\VEC{r}$) the resulting
optimization problem will be constructed from different sets of linear
or non-linear functions. We are interested in tracking the algebraic
properties of these functions as we construct mathematical programs
using the dynamics and constraints from
Eqs.~\eqref{eq:momentum_dynamics}, \eqref{eq:contact_constraints_2}.
Resulting objective functions and constraints will be classified as
elements from the following sets of functions.

\newtheorem{theorems:def}{Definition}
\begin{theorems:def}
  We define the set of affine functions
  \begin{align*}
    \mathcal{A} = \{ 
    &\VEC{s}: \mathbb{R}^n \rightarrow \mathbb{R}^m~ |~ \\
    &\exists \VEC{A} \in \mathbb{R}^{m \times n},~ \VEC{a} \in \mathbb{R}^m: \\
    &\VEC{s}(\VEC{x}) = \VEC{A} \VEC{x} + \VEC{a} ~\}
  \end{align*}
\end{theorems:def}
\begin{theorems:def}
  We define the set of positive semi definite quadratic functions
  \begin{align*}
    \mathcal{Q}_{+} = \{ 
    &\VEC{s}: \mathbb{R}^n \rightarrow \mathbb{R}^m~ |~\\
    &\exists \VEC{Q}_i \in \mathbb{R}^{n \times n} \text{ p.s.d},~
      \VEC{q}_i \in \mathbb{R}^n,~ c_i \in \mathbb{R}: \\
    &s_i(\VEC{x}) = \VEC{x}^T\VEC{Q}_i \VEC{x} + \VEC{q}_i\VEC{x} + c_i,~ i=1\dots m ~\}
  \end{align*}
\end{theorems:def}
\begin{theorems:def}
  We define the set of differences of positive semi definite functions
  \begin{align*}
    \mathcal{Q}_{\pm} = \{ 
    &\VEC{s}: \mathbb{R}^n \rightarrow \mathbb{R}^m~ |~\\
    &\exists \VEC{s'} \in \mathcal{Q}_{+},~
      \VEC{P}_i \in \mathbb{R}^{n \times n} \text{ p.s.d}: \\
    &s_i(\VEC{x}) = \VEC{s'}(\VEC{x}) - \VEC{x}^T\VEC{P}_i \VEC{x},~ i=1\dots m ~\}
  \end{align*}
  Throughout this paper we consider functions to be elements of
  $\mathcal{Q}_{\pm}$, only if the quadratic parameter matrices
  $\VEC{Q}_i,\VEC{P}_i$ can be accessed separately (instead of
  only having access to the indefinite matrix $\VEC{Q}_i -\VEC{P}_i$).
\end{theorems:def}

It is worth noting that $\mathcal{Q}_\pm$ is closed under addition,
scalar multiplication and composition with functions from
$\mathcal{A}$, i.e.

\begin{align}
  \label{eq:qpm_closed_under}
	\VEC{s}, \VEC{u} \in \mathcal{A};~ \VEC{v}, \VEC{w} \in \mathcal{Q}_\pm;
  \beta \in \mathbb{R} \nonumber \\ 
  \Rightarrow (\beta (\VEC{v} \circ \VEC{s}) + (\VEC{u} \circ \VEC{w}))
  \in \mathcal{Q}_\pm
\end{align}

Specifically, if we have quadratic parameter matrices $\VEC{Q}_i, \VEC{P}_i$
constructed separately then the operations in Eq.~\eqref{eq:qpm_closed_under}
preserve this separation.\\
As can be noticed from Eqs.~\eqref{eq:momentum_dynamics},
\eqref{eq:contact_constraints_2} the nonlinearity introduced in the
equations stems from cross products which can be classified as
follows
\newtheorem{theorems:theo}{Theorem}
\begin{theorems:theo} \label{theo:cross_product_is_dc} The function
  $\times(\VEC{a}, \VEC{b}): \mathbb{R}^6 \rightarrow \mathbb{R}^3,~
  \times(\VEC{a}, \VEC{b}) = \VEC{a} \times \VEC{b}$
  is an element of $\mathcal{Q}_{\pm}$ and we can write out the
  corresponding matrices $\VEC{Q}_i, \VEC{P}_i$ explicitly.
\end{theorems:theo}

\begin{proof}
It is straightforward to write a general cross product in a quadratic form with $[\VEC{a} \times \VEC{b}]_i = \begin{bmatrix}\VEC{a}^T & \VEC{b}^T\end{bmatrix}\VEC{H}_i\begin{bmatrix}\VEC{a}^T & \VEC{b}^T\end{bmatrix}^T, i=1,2,3$. Using an eigenvalue decomposition (for symmetric real matrices), we can then obtain $\VEC{H}_i = \VEC{V}_i (\VEC{D}_{+,i} - \VEC{D}_{-,i}) \VEC{V}_i^T$, with $\VEC{D}_{+,i}, \VEC{D}_{-,i}$ diagonal matrices with {\it only non-negative} values. Clearly, then $\VEC{a} \times \VEC{b} \in \mathcal{Q}_{\pm}$ with $\VEC{Q}_i = \VEC{V}_i \VEC{D}_{+,i} \VEC{V}_i^T, \VEC{P}_i = \VEC{V}_i \VEC{D}_{-,i} \VEC{V}_i^T$.
\end{proof} 

Note that constructing the quadratic parameter matrices $\VEC{Q}_i, \VEC{P}_i$ for a general cross
product can be carried out offline as it does not
require any problem data. Through out the construction in the remainder of this paper, we will not have to perform any eigenvalue decomposition as this  is typically costly. Instead, we exploit Eq.~\eqref{eq:qpm_closed_under} in order to construct functions from $\mathcal{Q}_{\pm}$.\\
It is straightforward to show that Eqs.~\eqsref{eq:contact_constraints_1}
{eq:contact_constraints_3} can be expressed as $\hat{\VEC{s}}(\hat{\VEC{f}},
\hat{\VEC{p}}, \hat{\tau}) \leq 0$, with $\hat{\VEC{s}} \in \mathcal{A}$,
i.e. if we choose to represent contact forces at $\hat{\VEC{p}}$ then the
contact constraints are affine. On the other hand, the angular momentum~
\eqref{eq:momentum_dynamics} will result in a non-linear function. In fact,
we will show that $\dVEC{k}(\hat{\VEC{f}}, \hat{\VEC{p}}, \hat{\tau})
\in \mathcal{Q}_{\pm}$ and we can construct the parameters $\VEC{Q}_i,
\VEC{P}_i$ separately inexpensively. On the other hand, if we chose to
represent the force torque at the CoM, we have the reverse effect, i.e. the
angular momentum becomes an affine function of the force-torques and as we will
show the contact constraints will be expressed as $\VEC{s}(\VEC{f},
\symVEC{\kappa}) \leq 0,~ \VEC{s} \in \mathcal{Q}_{\pm}$.\\

%
% Algebraic categorization of contact forces
%
\subsection{Decomposition of centroidal momentum dynamics}\label{sec:centroidal_dynamics_decomp}

In this section, we will categorize the nonlinear functions in
Eqs.~\eqref{eq:momentum_dynamics},~\eqsref{eq:contact_constraints_1}
{eq:contact_constraints_3} into one of the categories defined in
Sec.~\ref{sec:algebraic_categorization}. Categorizing the CoM torque
$\symVEC{\kappa}(\hat{\VEC{f}}, \hat{\tau}, \hat{\VEC{p}})$ (cf.
Eq.~\eqref{eq:com_torque_transformation}) requires consequent usage of
Eq.~\eqref{eq:qpm_closed_under} and
Theorem~\eqref{theo:cross_product_is_dc}. As a result one can show
that
$\symVEC{\kappa}(\hat{\VEC{f}}, \hat{\tau}, \hat{\VEC{p}}) \in
\mathcal{Q}_\pm$
and thus
$\dVEC{k}(\hat{\VEC{f}}, \hat{\tau}, \hat{\VEC{p}}) \in
\mathcal{Q}_\pm$.
Further, we will show that choosing to express external forces at the
CoM will result in contact constraints that are expressed as
inequalities on functions from $\mathcal{Q}_{\pm}$. Assuming a
unilateral contact, we can write the torque acting at the center of
pressure as
\begin{align*}
	\hat{\symVEC{\tau}} = \begin{bmatrix}0\\0\\\hat{\tau}\end{bmatrix}
&= \hat{\symVEC{\kappa}} + (\hat{\VEC{r}} - \begin{bmatrix}\hat{\VEC{p}} \\
	0\end{bmatrix}) \times \hat{\VEC{f}}\\
&= \VEC{R}^T\symVEC{\kappa}+ (\VEC{R}^T(\VEC{r} - \VEC{t}) -
  \begin{bmatrix}\hat{\VEC{p}} \\ 0\end{bmatrix}) \times (\VEC{R}^T\VEC{f})
\end{align*}
From the definition of the CoP we know that $\hat{\symVEC{\tau}}$
vanishes when projected on the contact surface, i.e. 

\begin{align}
    \VEC{S}^T &= \begin{bmatrix}1&0&0\\ 0&1&0\end{bmatrix}\nonumber \\
	\VEC{0} &= \VEC{S}^T\hat{\symVEC{\tau}}
  =  \VEC{S}^T((\VEC{R}^T(\VEC{r} - \VEC{t}))
    \times (\VEC{R}^T\VEC{f})) - \nonumber \\ 
    &~~~~~~~~~~~~~\VEC{S}^T( (\VEC{S}\hat{\VEC{p}}) \times (\VEC{R}^T\VEC{f})) + \VEC{S}^T\VEC{R}^T\symVEC{\kappa}\\
  \label{eq:p_hat_of_momentum}
  &= \VEC{R}_{x,y}^T \symVEC{\kappa} + \VEC{R}_{x,y}^T((\VEC{r} - \VEC{t}) \times \VEC{f}) + \VEC{S}^T [\VEC{R}^T\VEC{f}]_\times \VEC{S}\hat{\VEC{p}}
\end{align}
In order to resolve for $\hat{\VEC{p}}$, we first invert the premultiplied projector
\begin{align} \label{eq:p_projector}
	\VEC{S}^T [\VEC{R}^T\VEC{f}]_\times \VEC{S} = \VEC{R}_z^T \VEC{f}
	\begin{bmatrix}0& -1\\1& 0\end{bmatrix}\\
	\label{eq:p_projector_inv}
	(\VEC{S}^T [\VEC{R}^T\VEC{f}]_\times \VEC{S})^{-1} = \frac{1}{\VEC{R}_z^T \VEC{f}}
	\begin{bmatrix}0& 1\\-1& 0\end{bmatrix}
\end{align}
We note that from the friction constraint (cf. Eq.~\eqref{eq:contact_constraints_3}) we have
$\VEC{R}_z^T\VEC{f} > 0$. Next, we resolve
Eq.~\eqref{eq:p_hat_of_momentum} for $\hat{\VEC{p}}$ and substitute it
into the CoP constraints (cf. Eq.~\eqref{eq:contact_constraints_2}). For better readability we
will only show the result for the right-hand side of
Eq.~\eqref{eq:contact_constraints_2}
\begin{align}
	&~\VEC{p}_{max}  \geq \hat{\VEC{p}} - \VEC{c} \\
	\Leftrightarrow&~ \VEC{p}_{max} \geq-(\VEC{S}^T [\VEC{R}^T\VEC{f}]_\times \VEC{S})^{-1}
                   [ \VEC{R}_{x,y}^T \symVEC{\kappa} + \nonumber \\ 
                   &~ \VEC{R}_{x,y}^T((\VEC{r} -
                   \VEC{t}) \times \VEC{f} )] - \VEC{c}  \\
	\Leftrightarrow&~ ( \VEC{p}_{max} - \VEC{c}) \VEC{R}_z^T \VEC{f} \geq -\begin{bmatrix}0&1\\-1&0\end{bmatrix} ( \VEC{R}_{x,y}^T \symVEC{\kappa} + \nonumber \\%
&~ \VEC{R}_{x,y}^T((\VEC{r} - \VEC{t}) \times \VEC{f} )), \label{eq:nonlin_cop_constraint}
\end{align}
where we used Eq. \eqref{eq:p_projector_inv}. Applying Eq.~\eqref{eq:qpm_closed_under} and
Theorem~\eqref{theo:cross_product_is_dc}, one can transform
Eq.~\eqref{eq:nonlin_cop_constraint} into a function inequality
$\VEC{s}(\VEC{f}, \symVEC{\kappa}) \leq \VEC{0},~ \VEC{s} \in \mathcal{Q}_\pm$ and
again the quadratic parameter matrices can be constructed
inexpensively.

\section{Optimal Control Formulations}\label{sec:opt_ctrl_formulations}
Through out the following discussion, we will restrict ourselves to a predefined
contact activation pattern (cf. $\alpha_e$ in Eq.~\eqref{eq:momentum_dynamics}).
That means we predefine which end effector is in contact with the environment at
which time. This could for instance be computed with an acyclic contact planner \cite{tonneau:hal}. The location of the contact and the force-torque are expressed as decision variables and will be computed
by the optimizer. %
The goal of this section is to define an optimal control problem that finds
sequences of states and controls under the dynamics constraint in
Eq.~\eqref{eq:momentum_dynamics} and contact constraints
Eq.~\eqsref{eq:contact_constraints_1}{eq:contact_constraints_3} thus solving problem \eqref{eq:subprob_momentum}.

\subsection{Simultaneous Formulation}\label{sec:simultaneous_formulations}
We start out with a simultaneous formulation, i.e. we optimize over
both states and controls and express the dynamics constraint
explicitly in the mathematical program. Here we represent contact
forces at the center of pressure. We summarize the optimization variables into
$\VEC{x}_{0:T}=[\dots~ \hat{\VEC{f}}_{t,1:E}^T~ \hat{\VEC{p}}_{t,1:E}^T~
\hat{\symVEC{\tau}}_{t,1:E}^T~ \VEC{h}_{t+1}^T~ \dots]^T, t=0\dots T$ and $\VEC{c}_{1:E}$ with $1:E$ indexing all $E$ contacts, and we write our optimal control problem as 

\begin{align} \label{eq:qcqp_sim_1} \underset{\VEC{x}_{0:T}, \VEC{c}_{1:E}}
	{\text{min. }}& \sum_t^T J_t(\VEC{x}_t, \VEC{c}_{1:E}) \in \mathcal{Q}_+\\
  \label{eq:qcqp_sim_2}
	\text{s.t. }~ 1)~& \VEC{h}_{t+1} = \VEC{h}_t +
                \Delta \symVEC{f}_t( \hat{\VEC{f}}_{t,1:E}, \hat{\VEC{p}}_{t,1:E},
                \hat{\symVEC{\tau}}_{t,1:E} )  \\
  & \Leftrightarrow \VEC{0} = \VEC{x}_{t+1} + \symVEC{g}_t(\VEC{x}_t)\in \mathcal{Q}_\pm\\
  \label{eq:qcqp_sim_3}
  2)~&\text{Eqs.}~\eqsref{eq:contact_constraints_1}{eq:contact_constraints_3}\\
                & \Leftrightarrow \VEC{0} \leq \VEC{A}_t\VEC{x}_t + \VEC{A}_t'\VEC{c}_{1:E} \in \mathcal{A}
\end{align}
This formulation is a Quadratically constrained Quadratic Program (QCQP),
with a positive definite objective, linear inequality constraints
and nonlinear equality constraints on functions from $\mathcal{Q}_\pm$. As one can note from
\eqsref{eq:qcqp_sim_1}{eq:qcqp_sim_3}, summands of the objective as well as constraints
all depend only on time-local variables $\VEC{x}_{t:t+1}$ and on $\VEC{c}_{1:E}$. This
property has the advantage that optimization algorithms can be applied that
exploit the sparsity inside of objective Hessian and constraint Jacobian. 
The constraint Jacobian has block form, where we have a block tridiagonal matrix on the left and a dense matrix on the right. The number of columns in the dense part scale with $O(E)$, i.e. they do not grow as $T$ increases.
%
%The constraint Jacobian has
%the form
%
%\newcommand{\gDeriv}[1]{\frac{\partial \symVEC{g}_{#1}}{\partial \VEC{x}_{#1}}}
%
%\begin{equation} 
%\frac{\partial \VEC{g}}{\partial \VEC{x}_{0:T}, \VEC{c}_{q:E}} = 
%\begin{bmatrix} \frac{\partial \VEC{g}}{\partial \VEC{x}_{0:T}} & \frac{\partial \VEC{g}}{\partial \VEC{c}_{1:E} }    \end{bmatrix} = \left[\begin{array}{ccccc|c}
%*& * &    &   &   & * \\
% % & * & * &   &   & \vdots \\
% % &    &    & \dots &  &  \\
% % &    &    & * & * & * \\
%\end{array}\right]\\
%
%
%\left[\begin{array}{ccccc|c}
%\gDeriv{0} & \VEC{I}     &         &               &          &   \VEC{0} \\
%\VEC{A}_0   & \VEC{0}    &         &               &          &   \VEC{A}_0' \\
%
%            & \gDeriv{1} & \VEC{I} &               &          &   \VEC{0} \\
%            & \VEC{A}_1  & \VEC{0} &               &          &   \VEC{A}_1' \\
%
%            &            & \ddots  &               &          &   \vdots \\
%            &            &         &               &          &   \vdots \\
%
%            &            &         & \gDeriv{T-1}  & \VEC{I}  &   \VEC{0} \\
%            &            &         & \VEC{A}_{T-1}  & \VEC{0}  &   \VEC{A}_{T-1}' \\
%\end{array}\right]
%\end{equation}
%
%where we have a dense part on the right-hand side with number of columns in $O(E)$, i.e. they do not grow as $T$ increases. The top left-hand side is a band-diagonal structure. 
%
The Lagrangian $\mathcal{L}$ \cite{Nocedal2006NO} of the optimization program in Eq.~\eqref{eq:qcqp_sim_1} has a Hessian that can be decomposed into 4 blocks as follows

\newcommand{\lHessXX}[1]{ \nabla^2_{\VEC{x}_{#1}\VEC{x}_{#1}} ( J_{#1} + \sum \varphi_i \symVEC{g}_{#1,i} ) }
\newcommand{\lHessXC}[1]{ \nabla^2_{\VEC{x}_{#1}\VEC{c}_{1:E}}J_{#1} }
\newcommand{\lHessCC}[1]{ \nabla^2_{\VEC{c}_{1:E}\VEC{c}_{1:E}}J_{#1} }
\begin{eqnarray*} \label{eq:qcqp_sim_lagr_hess}
\left[\begin{array}{ccc|c}
\nabla_{\VEC{x}_0,\VEC{x}_0}^{2}\mathcal{L}  &            &      &   \\
%%%%&  &  \nabla_{\VEC{x}_1,\VEC{x}_1}^{2}\mathcal{L}   &            &      &   \\
      & \ddots &      &  \nabla_{\VEC{x},\VEC{c}}^2\mathcal{L}  \\
       &            &  \nabla_{\VEC{x}_T,\VEC{x}_T}^{2}\mathcal{L}   &   \\
\hline
&      \nabla_{\VEC{x},\VEC{c}}^{2}\mathcal{L}^T            &&  \nabla_{\VEC{c},\VEC{c}}^2\mathcal{L} \\
\end{array}\right]\\
%
%\begin{bmatrix}
%\lHessXX{0} &        &             & \lHessXC{0} \\
%
%            & \ddots &             & \vdots \\
%
%            &        & \lHessXX{T} & \lHessXC{T} \\
%
%\lHessXC{0} & \cdots & \lHessXC{T} & \lHessCC{T} \\
%\end{bmatrix}
\end{eqnarray*}

We note that again the dense blocks on the left and bottom do not depend on the time discretization $T$. The top-left block on the other hand is block-diagonal. This sparsity is favorable in optimization algorithms, because typically an essential step in these solvers is to solve a linear system that inherits the sparsity properties from the constraint Jacobian and Lagrange Hessian. Especially, interior-point methods are known to preserve the band-diagonal sparsity structure \cite{Dimitrov:2011bg}.

\subsection{Sparse Contact Force Representation} \label{sec:force_integral_representation}
In the following we will show that the sequential optimal control formulations for our problem can be transformed to an equivalent problem with a sparse structure similar to the simultaneous formulation. We will assume that a contact-phase $e$ is defined as the time interval $t \in [\sigma_e, \varepsilon_e)$ at which an end effector is touching the environment.  Thus the contact activation from Eq.~\eqref{eq:momentum_dynamics} can be refined to
\begin{align}
	\alpha_{e,t} = \begin{cases}
		1,& \text{if}~ \sigma_e \leq t < \varepsilon_e\\
		0,& \text{else}
	\end{cases}
\end{align}

where contact-phase $e$ starts at $t=\sigma_e$ and ends at $t=\varepsilon_e$. For instance, a contact-phase could describe a foot step $e$ where the heel hits the ground at $t=\sigma_e$ and the last time step at which the toe is touching the ground is $t=\varepsilon_e - 1$. Two heel strikes with the same foot would correspond to two different contact phases $e, e'$. Further, we define forces and CoM torques as linear functions of a new set of variables $\symVEC{\varphi}_t, \symVEC{\psi}_t$ as
\begin{align} \label{eq:force_intg_vars_1}
	&\VEC{f}_{e,t} = \symVEC{\varphi}_{e,t} - 2\symVEC{\varphi}_{e,t-1} + \symVEC{\varphi}_{e,t-2}\\
	&\dsymVEC{\kappa}_{e,t} = \symVEC{\psi}_{e,t} - 2\symVEC{\psi}_{e,t-1} + \symVEC{\psi}_{e,t-2}\\
	 \label{eq:force_intg_vars_3}
	&\forall t < \sigma_e: \symVEC{\varphi}_t = \symVEC{\psi}_t = 0
\end{align}

The variables $\symVEC{\varphi}_t, \symVEC{\psi}_t$ correspond to the twice integrated contact forces and torques. General optimal control problems that are expressed in sequential form, usually lose their sparsity structure because the state variables need to be integrated out. In the following, we will use the change of variables in Eqs.~\eqsref{eq:force_intg_vars_1}{eq:force_intg_vars_3} to show that this is not the case for our momentum optimization problem.
We will derive expressions for the state variables $\VEC{r}, \VEC{h}, \dVEC{h}$ as functions of $\symVEC{\varphi}_t, \symVEC{\psi}_t$. The goal here is to show that the state variables at  time step $t$ do \textit{not} depend on all previous time steps $t'=0 \dots t$, but only on a few ones. This has the advantage that a sequential formulation expressed with $\symVEC{\varphi}_t, \symVEC{\psi}_t$ will maintain its sparse structure and at the same time reduce the number of variables and constraints compared to the formulation in Sec.~\ref{sec:simultaneous_formulations}.\\
We compose the linear momentum rate as
\begin{align*}
	&\dBell_t(\symVEC{\varphi}_{e,t-2:t}) = 
	M\VEC{g} + \sum \dBell_{e,t}(\symVEC{\varphi}_{e,t-2:t}) \\
	&\dBell_{e,t} = \alpha_{e,t}\VEC{f}_{e,t} = \begin{cases}
		 \symVEC{\varphi}_{e,t} - 2\symVEC{\varphi}_{e,t-1} + \symVEC{\varphi}_{e,t-2},& \text{if } t<\varepsilon_e\\
		0,& \text{else}
	\end{cases} \\
\end{align*}
In order to construct expressions for the linear momentum and center of mass, we first define the sums of forces during a contact-phase over time
\begin{align*}
	\Bell_{e,t} &=  \sum_{i=0}^{t-1}\dBell_{e,i} = \begin{cases}
		 \symVEC{\varphi}_{e,t-1} - \symVEC{\varphi}_{e,t-2},& \text{if } t<\varepsilon_e\\
		 \symVEC{\varphi}_{e,\varepsilon_e-1} - \symVEC{\varphi}_{e,\varepsilon_e-2},& \text{else}
	\end{cases} \\
	M\VEC{r}_{e,t} &=  \sum_{i=0}^{t-1}\Bell_{e,i} = \begin{cases}
		 \symVEC{\varphi}_{e,t-2},& \text{if } t<\varepsilon_e\\
\parbox[t]{.35\linewidth}{$ \symVEC{\varphi}_{e,\varepsilon_e-2} + (t-\varepsilon_e) \times \\(\symVEC{\varphi}_{e,\varepsilon_e-1} - \symVEC{\varphi}_{e,\varepsilon_e-2})$ },& \text{else}
%		 \symVEC{\varphi}_{e,\varepsilon_e-2} + (t-\varepsilon_e)(\symVEC{\varphi}_{e,\varepsilon_e-1} - \symVEC{\varphi}_{e,\varepsilon_e-2}),& \text{else}
	\end{cases}
\end{align*}
With the derived expressions we can now construct the linear state terms as follows
\begin{align*}
	\dBell_t &= M\VEC{g} + \sum_e \dBell_{e,t}(\symVEC{\varphi}_{e,t-2:t})\\
	\Bell_t &= \Bell_0 + \Delta \sum_{i=0}^{t-1} \dBell \\
	&= \Bell_0 + \Delta tM\VEC{g} + \Delta^2 \sum_e \Bell_{e,t}(
	\symVEC{\varphi}_{e,t-2:t}, \symVEC{\varphi}_{e,\varepsilon_e-2:\varepsilon_e-1})\\
	M\VEC{r}_t &= M\VEC{r}_0 + \Delta \sum_{i=0}^{t-1}\Bell_t = M\VEC{r}_0 + \Delta t\Bell_0 + \Delta^2  \frac{t(t-1)}{2}M\VEC{g} + \\
	&+ \Delta^2 \sum_e M\VEC{r}_{e,t}(
	 \symVEC{\varphi}_{e,t-2}, \symVEC{\varphi}_{e,\varepsilon_e-2:\varepsilon_e-1}),
\end{align*}

where $\Delta$ is a discretization time.
It is interesting to see, that the state variables at time $t$ only depend on
control variables at $t' \in \{t-2, t-1, \varepsilon_e-1, \varepsilon_e-2|
e=1\dots E\}$. This has the advantage that derivatives of the objective function
and constraints remain sparse and as such result in more efficient optimization
algorithms similar to the simultaneous formulation. In the previous derivation we obtained expressions for the linear momentum, but expressions 
for the angular momentum can be readily derived with the same sparsity properties.

\subsection{Sparse Sequential Optimal Control}\label{sec:sparse_seq_formulation}

We will now express a sequential optimal control problem with respect to integrals
of forces as they were defined in Sec.~\ref{sec:force_integral_representation}. We
concatenate decision variables into vectors $\VEC{x}_{0:T} = [\dots~ \symVEC{\varphi}^T_{t,1:E}~ \symVEC{\psi}^T_{t,1:E}~ \dots]^T,
\VEC{y}_{1:E} = [\VEC{c}^T_{1:E}~ \symVEC{\varphi}^T_{\varepsilon_e-2:\varepsilon_e,1:E}~ \symVEC{\psi}^T_{\varepsilon_e-2:\varepsilon_e,1:E}]^T$

\begin{align} \label{eq:qcqp_seq_1} \underset{\VEC{x}_{0:T}, \VEC{y}_{1:E}}
	{\text{min. }}& \sum_t^T J_t(\VEC{x}_{t-2:t}, \VEC{y}_{1:E}) \in \mathcal{Q}_+\\
  \label{eq:qcqp_seq_2}
	\text{s.t. }~ 1)~& \text{Eq.}~\eqref{eq:nonlin_cop_constraint}\\
  & \Leftrightarrow \VEC{0} \leq \symVEC{g}_t(\VEC{x}_{t-2:t}, \VEC{y}_{1:E}) \in \mathcal{Q}_\pm\\
  \label{eq:qcqp_seq_3}
  2)~&\text{Eqs.}~\eqref{eq:contact_constraints_3}\\
                & \Leftrightarrow \VEC{0} \leq \VEC{A}_t\VEC{x}_{t-2:t} \in \mathcal{A}
\end{align}

We can see that similar to the simultaneous formulation, summands of the objective
and constraint functions only depend on $\VEC{x}_{t-2:t}, \VEC{y}_{1:E}$ rather than
on all time steps. The quadratic constraints in Eq.~\eqref{eq:qcqp_seq_2} can be decomposed efficiently as discussed in Sec.~\ref{sec:centroidal_dynamics_decomp}. We will now write out the sparse constraint Jacobian

\newcommand{\gDerivX}[2]{\frac{\partial \symVEC{g}_{#1}}{\partial \VEC{x}_{#2}}}
\newcommand{\gDerivY}[1]{\frac{\partial \symVEC{g}_{#1}}{\partial \VEC{y}_{1:E}}}
\begin{equation} 
\left[\begin{array}{cccccc|c}
\gDerivX{0}{0} & \gDerivX{0}{1} &\gDerivX{0}{2}  &               &                                       &                    & \gDerivY{0} \\
\VEC{A}_0      & \VEC{0}        &\VEC{0}         &               &                                       &                    & \VEC{0}     \\
               & \gDerivX{1}{1} & \gDerivX{1}{2} &\gDerivX{1}{3} &                                       &                    & \gDerivY{1} \\
               & \VEC{A}_1      & \VEC{0}        &\VEC{0}        &                                       &                    & \VEC{0}     \\
               &                & \ddots         &                                  &                    &                    &   \vdots \\
               &                &                &                                  &                    &                    &   \vdots \\
               &                &                &               \gDerivX{T-1}{T-3} & \gDerivX{T-1}{T-2} &\gDerivX{T-1}{T-1}  & \gDerivY{T-1} \\
               &                &                &                \VEC{A}_{T-1}      & \VEC{0}            &\VEC{0}             & \VEC{0}  
\end{array}\right] \nonumber 
\end{equation}

We will now write out the Hessian of the Lagrangian

\renewcommand{\lHessXX}[2]{ \nabla^2_{\VEC{x}_{#1}\VEC{x}_{#2}} \mathcal{L} }
\newcommand{\lHessXY}[1]{ \nabla^2_{\VEC{x}_{#1}\VEC{y}_{1:E}} \mathcal{L} }
\newcommand{\lHessYY}{ \nabla^2_{\VEC{y}_{1:E}\VEC{y}_{1:E}} \mathcal{L} }
%
%\begin{equation} \label{eq:qcqp_sim_lagr_hess}
%\begin{bmatrix}
%\lHessXX{0}{0} & \lHessXX{0}{1} & \lHessXX{0}{2} &                  &                  &                  & \lHessXY{0} \\
%
%\lHessXX{0}{1} & \lHessXX{1}{1} & \lHessXX{1}{2} & \lHessXX{1}{3}   &                  &                  & \lHessXY{1} \\
%
%\lHessXX{0}{2} & \lHessXX{1}{2} & \ddots         & \ddots           & \ddots           &                  & \vdots      \\
%
%               & \lHessXX{1}{3} & \ddots         &                  &                  & \lHessXX{T-2}{T} &             \\
%
%               &                & \ddots         &                  &                  & \lHessXX{T-1}{T} &             \\
%
%               &                &                & \lHessXX{T-2}{T} & \lHessXX{T-1}{T} & \lHessXX{T}{T}   &             \\
%
%\lHessXY{0}    & \lHessXY{1}    & \cdots         &                  &                  &                  & \lHessYY    \\
%\end{bmatrix}
%\end{equation}
%
\begin{eqnarray}
\left[\begin{array}{ccccc|c}
\nabla_{0,0}^2& \nabla_{0,1}^2 &  \multicolumn{1}{c|}{\nabla_{0,2}^2 } &   &   &  \\ \cline{4-4}
\nabla_{0,1}^2& \nabla_{1,1}^2 &  & \multicolumn{1}{c|}{ } &   &  \\ \cline{5-5}
\nabla_{0,2}^2&  & \ddots &   & \nabla_{T,T-2}^2	&  \nabla_{\VEC{x},\VEC{y}}^2\mathcal{L}\\ \cline{1-1}
 &  \multicolumn{1}{|c}{ } &  &  &  \nabla_{T,T-1}^2   & \\ \cline{2-2}
 &   &  \multicolumn{1}{|c}{ \nabla_{T,T-2}^2} &  \nabla_{T,T-1}^2&  \nabla_{T,T}^2    &  \\
\hline 
\multicolumn{5}{c|}{\nabla_{\VEC{x},\VEC{y}}^2\mathcal{L}^T} &  \nabla_{\VEC{y},\VEC{y}}^2\mathcal{L}\\
\end{array}\right] \nonumber 
\end{eqnarray}
\begin{align}
  \nabla^2_{t,t'} = \nabla^2_{\VEC{x}_t\VEC{x}_{t'}}\mathcal{L} &=
  %\nabla^2_{\VEC{x}_t\VEC{x}_{t'}}(J_t+J_{t+1}+J_{t+2} + \\
  %
  %\sum_i\phi_i (
  %\symVEC{g}_{t,i}+\symVEC{g}_{t+1,i}+\symVEC{g}_{t+2,i}) ) \\
  \sum_{k=0}^2 \nabla_{t,t'}^2 J_{t+k} + \sum_{k=0}^2 \phi_{t+k,i} \nabla_{t,t'}^2 \VEC{g}_{t+k, i}, \nonumber \\
  %
  %= \sum_{k=0}^2 \nabla_{t,t'}^2 J_{t+k} + \sum_{k=0}^2 \phi_{t+k,i} (\VEC{Q}_{t+k, i} - \VEC{P}_{t+k, i}) \in \mathcal{Q}_\pm \nonumber  
  %
  \label{eq:lagr_decomp}
  \nabla_{t,t'}^2 \VEC{g}_{t+k, i}  &= (\VEC{Q}_{t+k, i} - \VEC{P}_{t+k, i}) \in \mathcal{Q}_\pm
\end{align}

\subsection{Comparison of sparsity patterns}\label{sec:comparison_sparsity}
As we analyzed in the previous sections, both the simultaneous as well as sequential formulation of the mathematical program ~\eqref{eq:subprob_momentum} have sparse constraint Jacobians and Lagrangian Hessians. A common approach to solve optimal control formulations of that kind is to use interior point methods \cite{Domahidi:2012gs} as computation time benefits from these sparsity properties. Usually, the computation time is dominated by a decomposition of an {\it augmented system} matrix $\nabla^2\tilde{\mathcal{L}} + \VEC{G}^T\symVEC{\Sigma}\VEC{G}$, where $\VEC{G}$ is the constraint Jacobian and $\nabla^2\tilde{\mathcal{L}}$ is a (preferably convex) approximation to the Lagrangian Hessian. Comparing the sparsity patterns of the two formulations from Sec. \ref{sec:opt_ctrl_formulations}, we see that both problems result in augmented systems that have a block structure with a band-diagonal matrix on the top left. The computational complexity for decomposing the augmented system is in both formulations linear w.r.t. the discretization parameter $T$.
Finding an approximation $\nabla^2\tilde{\mathcal{L}}$ that both makes progress towards a solution and can be robustly decomposed is usually not trivial. In our problem however, a convex approximation can be constructed exploiting the fact that $\nabla^2\mathcal{L}$ can be decomposed into a difference of p.s.d matrices (cf. Eq. \eqref{eq:lagr_decomp}). We can obtain $\nabla^2\tilde{\mathcal{L}}$ by simply dropping the negative part of $\nabla^2\mathcal{L}$ from Eq. \eqref{eq:lagr_decomp}. Note that we do not require expensive operations such as eigenvalue decompositions to be carried out online, but obtain the convex part of $\nabla^2\mathcal{L}$ through consequently applying Eq.~\eqref{eq:qpm_closed_under} and
Theorem~\eqref{theo:cross_product_is_dc} as we construct the problem. This can be done for both, the simultaneous as well as the sequential formulation.\\
The number of variables is very different for the sequential and simultaneous formulations and has an impact on the computational complexity of resulting optimization algorithms. In the simultaneous case, we have to consider 6 force variables for each end effector and 9 state variables. For instance a biped that remains in double support at all time requires $21T$ variables. On the other hand, the sequential method does not require variables for states and thus contains only $12T$ variables for the aforementioned example. Further, we drop the $6T$ dynamics constraints. As a consequence computational complexity is reduced with a sequential formulation as we solve for longer and finer grained time horizons $T$. \\
It is also worth mentioning that since the momentum dynamics generalize the commonly used LIPM, the sparse structure in our sequential formulation will be similar using LIPM dynamics. This nicely complements the discussion in \cite{Dimitrov:2011bg}.

\section{Experiments}\label{sec:experiments}
We demonstrate the proposed motion optimization Algorithm ~\ref{algo:motion_generation} on a stepping task for a humanoid robot. Further, we convince ourselves that the sparse structure in our momentum problem results in linear computational complexity w.r.t. time discretization $T$. All experiments were performed on a notebook with a 2.7 GHz intel i7 processor with 16gb RAM. 

\begin{figure}
  \centering
   \subfigure{\begin{minipage}{0.23\textwidth}
      \includegraphics[trim=0 0 0 65, clip, width=\linewidth]{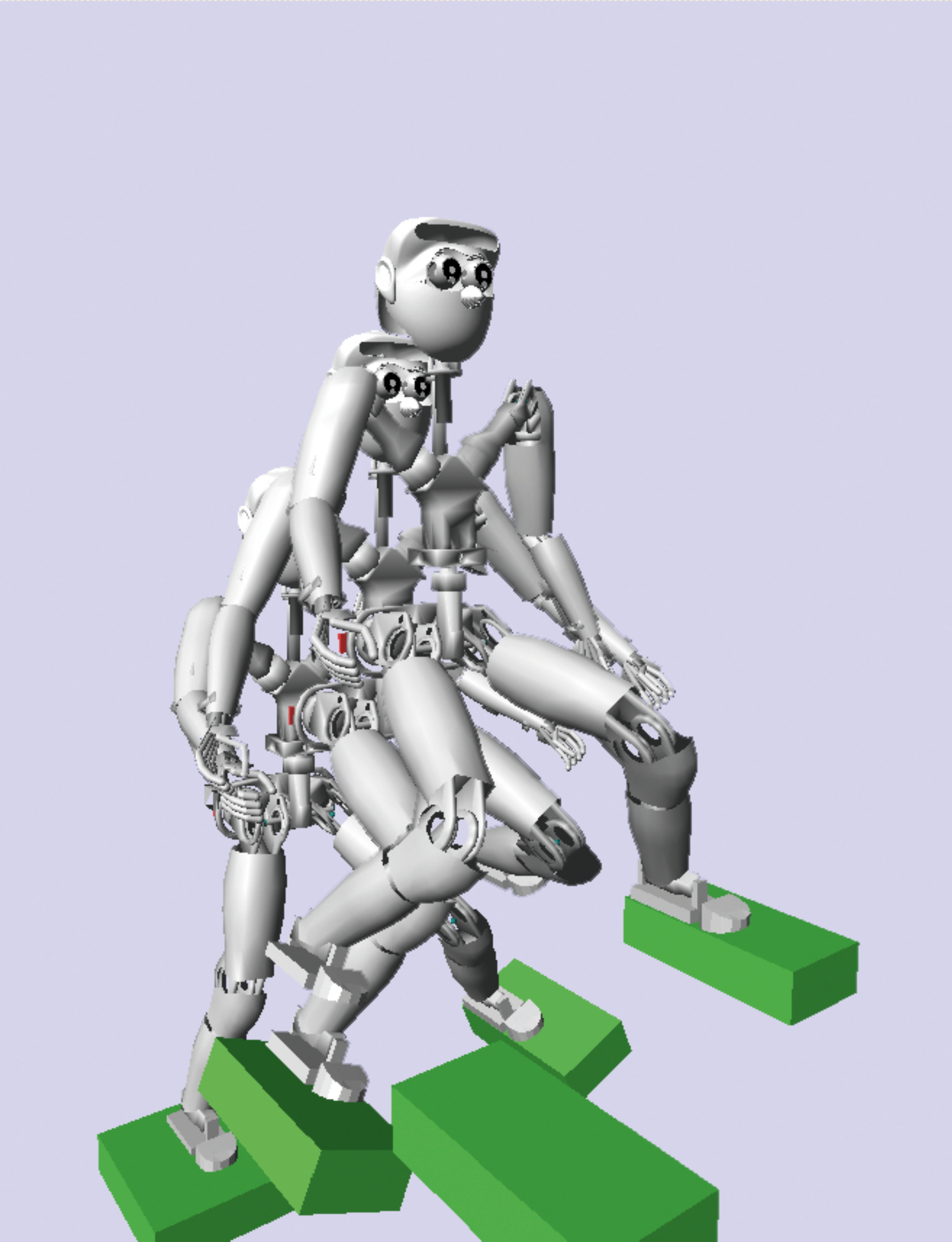}
    \end{minipage}} 
    \subfigure{\begin{minipage}{0.23\textwidth}
     %\vspace{11pt}
      \includegraphics[width=\linewidth]{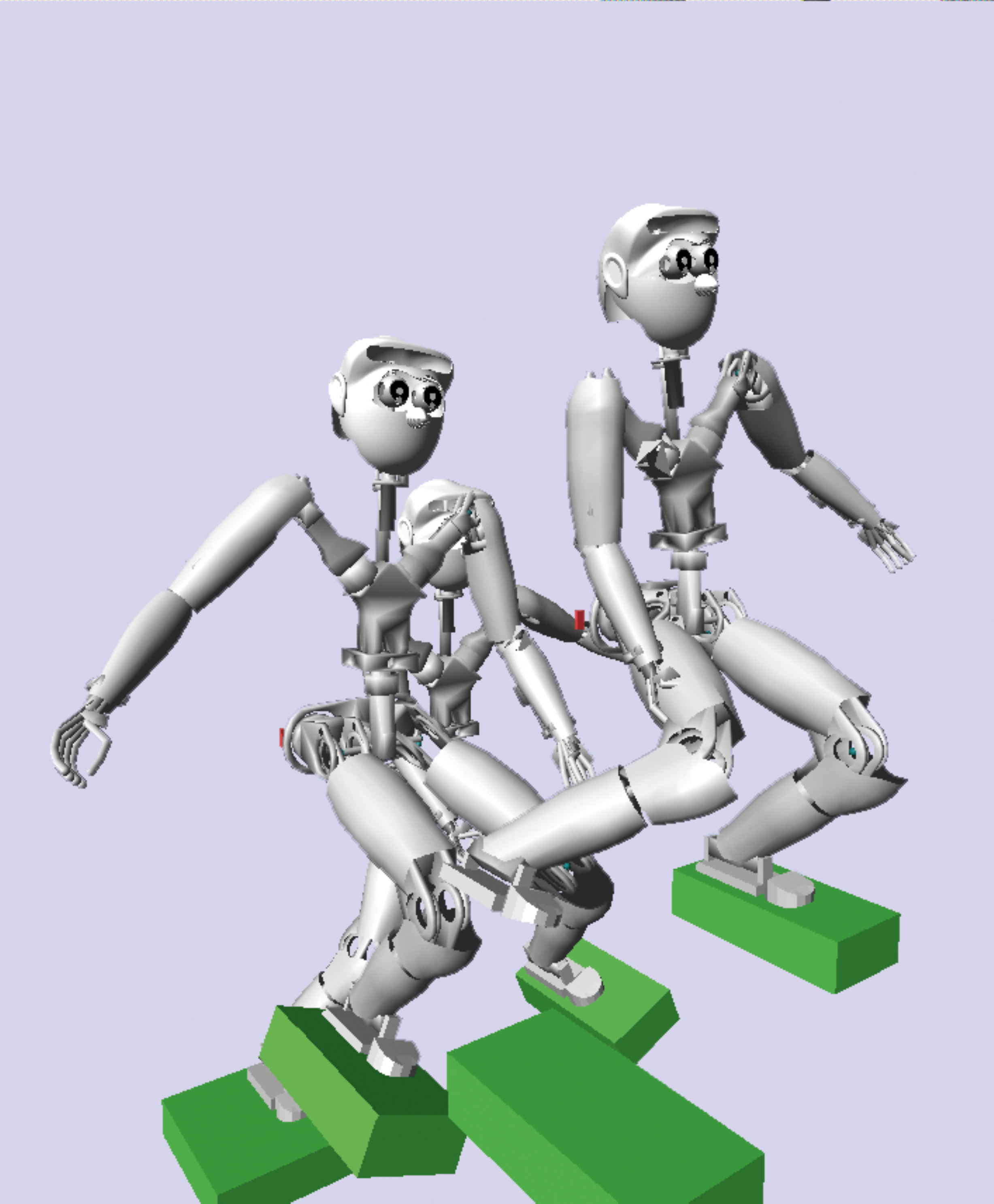}
    \end{minipage}}
  \vspace{-.3cm}
  \caption{\scriptsize A visualization of a kinematic plan generated by Algorithm \ref{algo:motion_generation}. After the first iteration (left) the CoM remains above the center between the feet and the arms are at rest. After the last iteration (right) the kinematic plan is in coherence with constrained force profiles. The CoM has a wider sway in lateral direction and the robot uses the arms to account for angular momentum.}\label{fig:gait_motion}
  \vspace{-0.4cm}
\end{figure}

\subsection{Optimization Algorithm}

\begin{figure}
  \centering
  \includegraphics[width=\linewidth]{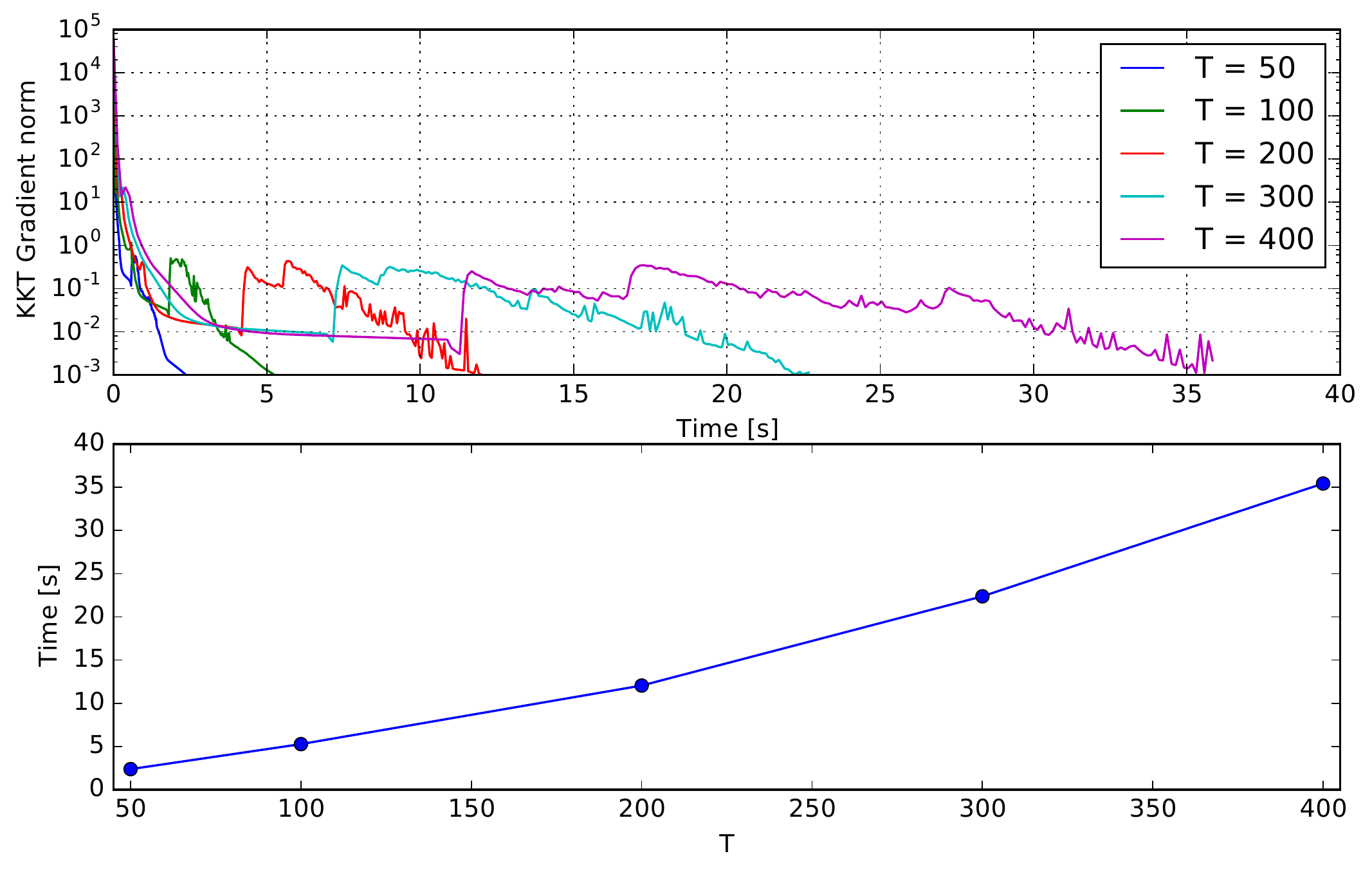}
  \vspace{-.7cm}
  \caption{\scriptsize Computation time of the momentum optimization \eqref{eq:subprob_momentum}. We execute the optimization for increasing values T and plot the resulting norm of the Lagrangian gradient (top) and the overall computation time (bottom). Although our implementation still requires improvements, we can already see a linear trend in the computation complexity confirming our analysis in Sec. \ref{sec:sparse_seq_formulation}}\label{fig:computation_time}
  \vspace{-0.2cm} 
\end{figure}

In this section we would like to discuss our choice of optimization algorithms for solving the two subproblems in \eqref{eq:subprob_momentum},\eqref{eq:subproblem_motion}. In both algorithms, we exploit the sparse structure resulting from the optimal control formulation. Further, we condense the number of variables in the momentum optimal control problem \eqref{eq:subprob_momentum} as we discussed in Sec. ~\ref{sec:sparse_seq_formulation}. 
We solve the optimization problem \eqref{eq:subproblem_motion} with a Gauss-Newton method, where the Hessian is a block tridiagonal positive definite matrix that can be factorized with a dedicated Cholesky Decomposition \cite{Wang:2010}. The bottleneck in our implementation is a numerical differentiation of the momentum Jacobian $\frac{\partial \VEC{h}(\VEC{q}, \dVEC{q}) }{\partial \VEC{q}}$, which should be constructed directly from the dynamics model. Nevertheless, with our naive implementation we were able to generate whole-body motion and force plans in 30 s of computation time. \\
For the momentum optimization \eqref{eq:subprob_momentum} we implement a primal-dual interior point method \cite[Algo.~19.2]{Nocedal2006NO}, which allows us to preserve the band-diagonal structure in our problem. The algorithm is modified as discussed in Sec \ref{sec:comparison_sparsity} to exploit the convex approximation of the Lagrangian Hessian. We run the momentum optimization for increasing values of $T$ and plot convergence of the Lagrangian gradient in Fig. \ref{fig:computation_time}. It becomes evident from the plot that computation time increases linearly as we increase $T$. Note that sequential methods in general scale cubic with $T$, however, our sparse formulation allows to preserve linear complexity potentially leading to more efficient algorithms.\\
In a side experiment we investigated the effect of our approximation $\nabla^2\tilde{\mathcal{L}}$ on convergence to a solution. We implemented a very basic sequential quadratic program \cite[Algo.~18.1]{Nocedal2006NO} that was using dense operations, meaning it was not exploiting the sparsity structure. Instead we focused on the benefits of the approximation $\nabla^2\tilde{\mathcal{L}}$. Although much slower than the sparse interior-point method, our algorithm still converged by an order of magnitude faster than an off-the-shelf SQP method, SNOPT, \cite{Gill:2002} motivating the need for dedicated solvers in robotics.
\begin{figure}
  \centering
  \includegraphics[width=\linewidth]{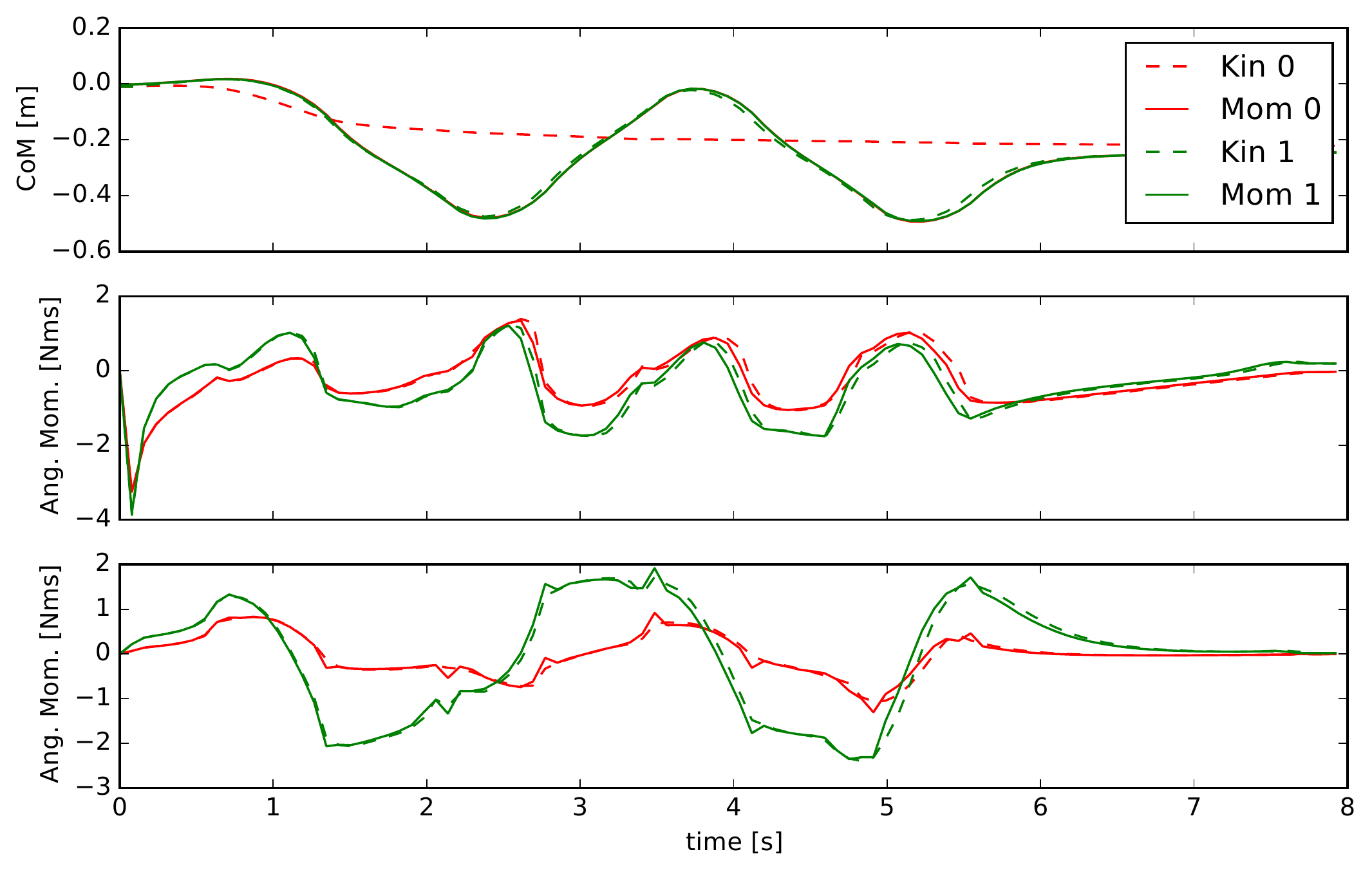}
  \vspace{-.7cm}
  \caption{\scriptsize Momentum plans generated by the motion optimization algorithm \ref{algo:motion_generation}. The kinematic pass optimizes for an unfeasible CoM lateral motion, which is corrected after the first momentum optimization. In the second iteration of the algorithm, the kinematic plan is adapted to realize the CoM sway. Horizontal angular momentum (bottom two plots) is also found to be consistent across kinematic as well as momentum optimization.}\label{fig:plan}
  \vspace{-0.4cm}
\end{figure}

\subsection{Motion Generation}

\begin{figure}
  \centering
  \includegraphics[width=.5\linewidth]{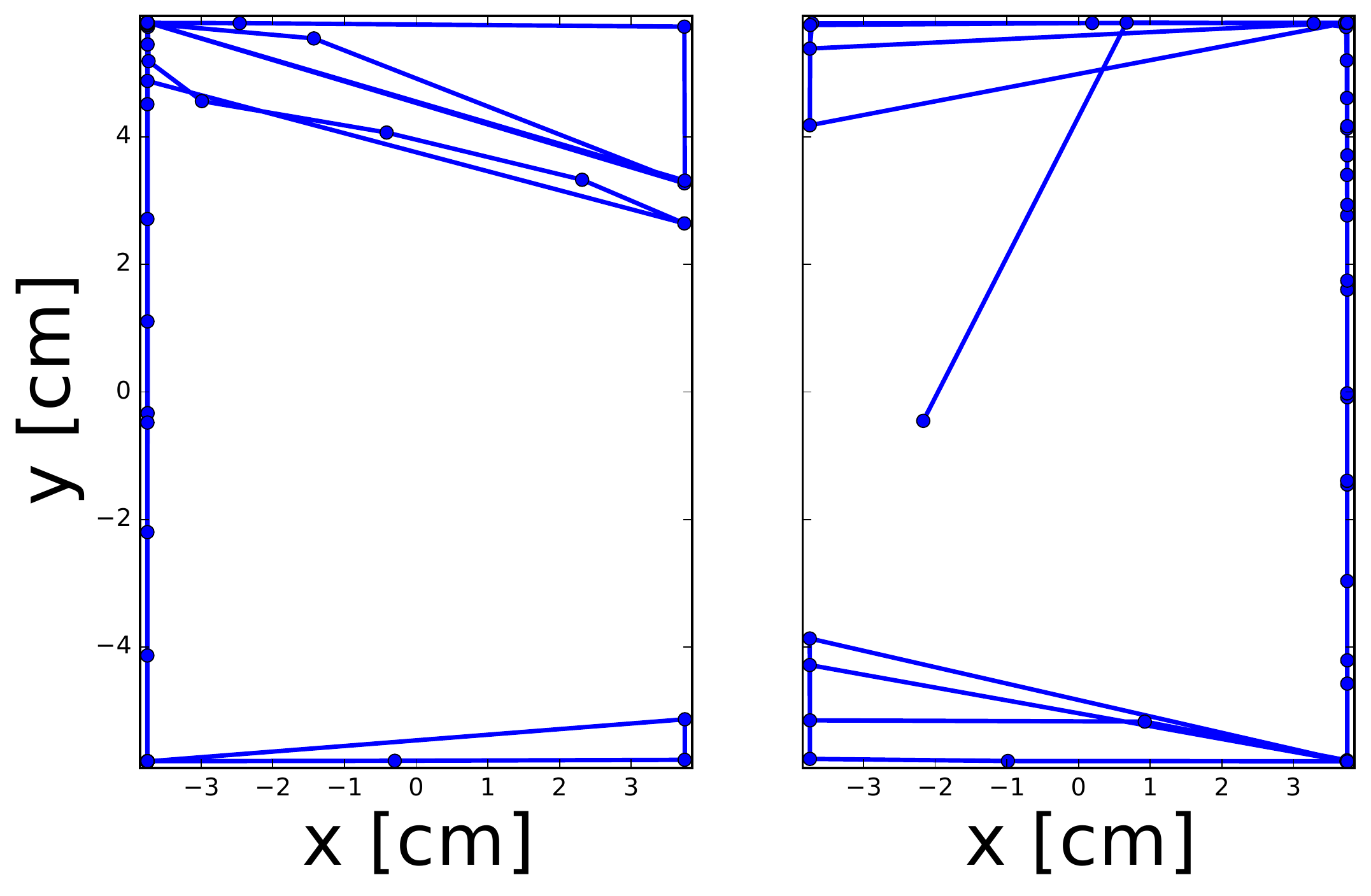}
  \vspace{-.2cm}
  \caption{\scriptsize Footprints of the robot with optimized CoP trajectories. As can be seen from the plot, the CoPs never exceed the foot support.}\label{fig:footprints}
  \vspace{-0.5cm}
\end{figure}

We construct a stepping scenario\footnote{The result is summarized in \url{https://youtu.be/NkP5Z9MfRRw}} for a simulation of our Sarcos humanoid robot (cf. Fig. \ref{fig:gait_motion}). The humanoid is to step on stepping stones that increase in height and are angled as illustrated in the figure. Note that here the LIPM assumption does not hold, but instead non-coplanar contacts have to be considered together with angular momentum that is required for leg swing motions. In our experiments, we fix the footstep locations $\VEC{c}_e$ to predefined values for a simpler implementation and compute coherent joint trajectories and force profiles using Algorithm \ref{algo:motion_generation}. Every $0.7$s contact is broken or created switching from single to double support or vice versa. Our initial guess for the CoM and momentum $\bar{\VEC{h}}$ is to simply have zero momentum and keep the CoM centered above the feet. We introduce weights into the cost function in Eq.~\eqref{eq:subproblem_motion} giving more importance to footstep locations. In our example, Algorithm \ref{algo:motion_generation} converges already after 30s for $T=100$, despite the inefficient implementation and requires only 2 passes to acquire consistent force and joint trajectories. The first kinematic plan realizes our naive CoM motion moderately well, which however cannot be accomplished given the contact constraints. Thus, the momentum optimization introduce a CoM sway in lateral direction, which is then realized by the kinematic optimizer in the consecutive iteration as can be seen in Fig.~\ref{fig:plan}. Additionally, the kinematic plan introduces an arm sway motion after the first pass to compensate for angular momentum introduced by the swing legs. After the motion generator converges it yields a kinematic plan that is coherent with bounded force and CoP profiles (cf. Fig. \ref{fig:footprints}). In this example, we keep the ankle joints fixed and do not account for object collision. However, collision avoidance can be added inside the kinematic optimization using trajectory optimization techniques \cite{Kalakrishnan2011}.\\
Our experiments demonstrate that constrained force profiles and consistent whole-body motion plans can be generated efficiently with the proposed algorithm. The structure identified in the momentum optimization led to more efficient optimization algorithms.

\section{Conclusion}
We presented a contact-centric motion generation algorithm for floating-base robots. It consists of two sub-problems optimizing for kinematic trajectories and momentum profiles iteratively. An analysis of the momentum optimization is presented and the relevance for optimal control formulations is discussed. In our experiments, we demonstrate the momentum-centric motion generator on a stepping task for a humanoid robot simulation. Indeed, consistent momentum and whole-body motion plans can be acquired efficiently. Further, our experiments show favorable scaling properties of our sparse optimal control formulation.

%
%\section*{Acknowledgments}
%
%
\bibliographystyle{IEEEtran}
{\footnotesize \bibliography{iros2016} }
\end{document}